\documentclass[]{bytedance}
\usepackage[toc,page,header]{appendix}

\usepackage{minitoc}
\usepackage{amsfonts}
\usepackage{amssymb}
\usepackage{tabularx}
\usepackage{listings}
\usepackage{xcolor}
\usepackage{cancel}

\usepackage{tabulary,multirow,xspace}
\usepackage{fixmath,mathtools,nicefrac,mmstyle}
\usepackage{subcaption}
\usepackage{caption}
\usepackage{float}
\usepackage{wrapfig} % for wrapfigure
\usepackage[misc]{ifsym} % to use \Letter
\usepackage{colortbl}

\usepackage{wrapfig}
\usepackage{multicol}
\usepackage[most]{tcolorbox}
\usepackage{pifont}

\definecolor{codegreen}{rgb}{0,0.6,0}
\definecolor{codegray}{rgb}{0.5,0.5,0.5}
\definecolor{codepurple}{rgb}{0.58,0,0.82}
\definecolor{backcolour}{rgb}{0.95,0.95,0.92}
\definecolor{boxblue}{RGB}{57,89,163}
\definecolor{boxbluebg}{RGB}{230,237,250} 

\lstdefinestyle{mystyle}{
    backgroundcolor=\color{backcolour},   
    commentstyle=\color{codegreen},
    keywordstyle=\color{magenta},
    numberstyle=\tiny\color{codegray},
    stringstyle=\color{codepurple},
    basicstyle=\ttfamily\footnotesize,
    breakatwhitespace=false,         
    breaklines=true,                 
    captionpos=b,                    
    keepspaces=true,                 
    numbers=none,                    
    numbersep=5pt,                  
    showspaces=false,                
    showstringspaces=false,
    showtabs=false,                  
    tabsize=2
}
\definecolor{mygray1}{gray}{.95}
\definecolor{mygray2}{gray}{.9}
\definecolor{mygray3}{gray}{.95}
\usepackage{pifont}% http://ctan.org/pkg/pifont
\newlength\savewidth
\newcolumntype{x}[1]{>{\centering\arraybackslash}p{#1pt}}

\newcommand{\app}{\raise.17ex\hbox{$\scriptstyle\sim$}}

\usepackage{xcolor}
\usepackage{listings}

\definecolor{commentgreen}{rgb}{0.1, 0.4, 0.1}
\definecolor{keywordblue}{rgb}{0.1, 0.1, 0.7}
\definecolor{stringred}{rgb}{0.7, 0.1, 0.1}

\lstdefinestyle{mystyle}{
    commentstyle=\color{commentgreen},
    keywordstyle=\color{keywordblue},   
    stringstyle=\color{stringred},
    basicstyle=\ttfamily\scriptsize, 
    breaklines=true,
    keepspaces=true,
    showstringspaces=false,
    frame=none,                     
    language=Python, 
}

\newcommand{\name}{CyberV}
\title{\name{}: Cybernetics for Test-time Scaling in Video Understanding}

\author{
\centerline{
Jiahao Meng\textsuperscript{\rm 1} \qquad
Shuyang Sun\textsuperscript{\rm 2}  \qquad
Yue Tan\textsuperscript{\rm 1}  \qquad
Lu Qi\textsuperscript{\rm 2}  \qquad
Yunhai Tong\textsuperscript{\rm 1}  \qquad
} 
\centerline{
Xiangtai Li\textsuperscript{\rm 2}\textsuperscript{$\dagger$} \qquad 
Longyin Wen\textsuperscript{\rm 2} \qquad
}
}
\affiliation[]{\textsuperscript{\rm 1} Peking University \quad
\textsuperscript{\rm 2} ByteDance}
\abstract{
Current Multimodal Large Language Models (MLLMs) may struggle with understanding long or complex videos due to computational demands at test time, lack of robustness, and limited accuracy, primarily stemming from their feed-forward processing nature. These limitations could be more severe for models with fewer parameters.
To address these limitations, we propose a novel framework inspired by cybernetic principles, redesigning video MLLMs as adaptive systems capable of self-monitoring, self-correction, and dynamic resource allocation during inference.
Our approach, \textbf{\name{}}, introduces a cybernetic loop consisting of an MLLM Inference System, a Sensor, and a Controller. Specifically, the sensor monitors forward processes of the MLLM and collects intermediate interpretations, such as attention drift, then the controller determines when and how to trigger self-correction and generate feedback to guide the next round.
This test-time adaptive scaling framework enhances frozen MLLMs without requiring retraining or additional components. 
Experiments demonstrate significant improvements: \textbf{\name{}} boosts Qwen2.5-VL-7B by 8.3\% and InternVL3-8B by 5.5\% on VideoMMMU, surpassing the competitive proprietary model GPT-4o. When applied to Qwen2.5-VL-72B, it yields a 10.0\% improvement, achieving performance even comparable to human experts. Furthermore, our method demonstrates consistent gains on general-purpose benchmarks, such as VideoMME and WorldSense, highlighting its effectiveness and generalization capabilities in making MLLMs more robust and accurate for dynamic video understanding.
The code is released at \url{https://github.com/marinero4972/CyberV}.
}

\date{\today}

\correspondence{Yunhai Tong at \email{yhtong@pku.edu.cn}, Xiangtai Li at \email{xiangtai.li@bytedance.com}}

\begin{document}
\maketitle

\begin{figure}[h]
\centering
\includegraphics[width=0.94\linewidth]{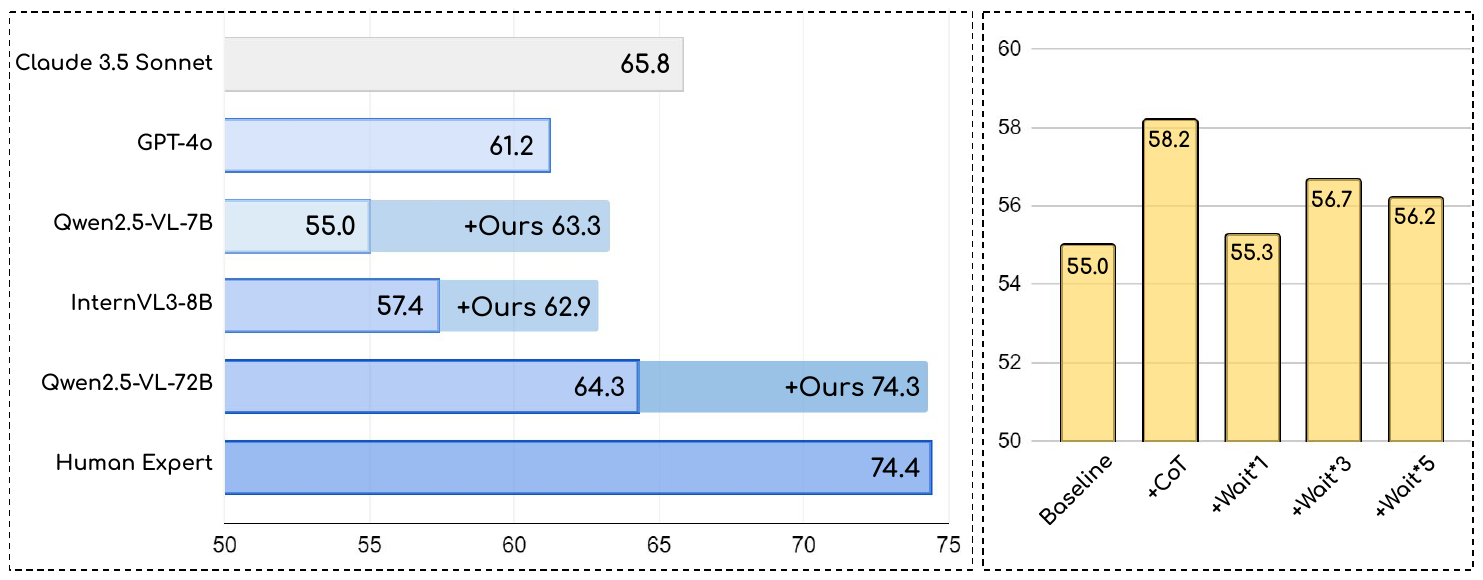}
\caption{\small{Performance comparison on the VideoMMMU benchmark.
\textbf{Left:} \name{} boosts small open-source models with only 7B parameters to outperform GPT-4o; with a larger model, \name{} surpasses the prior state-of-the-art and approaches the human level.
\textbf{Right:} CoT reasoning improves results when using Qwen2.5-VL-7B, but multi-round reflection via ``Wait'' degrades performance.}}
\label{fig:teaser}
\end{figure}

\section{Introduction}
\label{sec:intro}

\vspace{-2pt}

Understanding dynamic visual scenes in videos is a fundamental challenge, with applications ranging from autonomous driving to content analysis and human-robot interaction. 
Multimodal Large Language Models (MLLMs) have recently emerged as a powerful paradigm, demonstrating impressive capabilities by integrating pre-trained large language models with visual encoders to process and reason about video content~\cite{2023visionllm,bai2023qwen,team2024gemini,llava-video}. 
However, deploying these models effectively, particularly for understanding long, complex, or nuanced videos at test time, presents significant hurdles. 
Current MLLMs often struggle with the computational demands of processing extended video streams (test-time scaling), exhibit brittleness to variations or unexpected events in the input (lack of robustness), and are prone to generating inaccurate, inconsistent, or hallucinatory interpretations (limited accuracy)~\cite{llava-video, li2024llavaonevision,video-of-thought,hu2025cos,han2024free}. 
In particular, as shown in Figure~\ref{fig:teaser}, directly adding the reflection prompts to the output during inference does not perform well in video understanding. Even the current state-of-the-art MLLM cannot effectively scale up its capabilities.

We argue that these limitations stem from the feed-forward processing pipeline inherent in current MLLM architectures.  
%predominantly
%
These models typically process videos in a single, often computationally intensive, pass, lacking mechanisms for dynamic adaptation, self-correction, or targeted analysis based on evolving understanding or specific task demands. 
This contrasts sharply with biological systems, which continuously use feedback to regulate behavior and adapt to complex environments.

To address this gap, we propose incorporating principles from cybernetics – the study of control, communication, and self-regulation in systems \cite{wiener1948cybernetics, ashby1956introduction} – into the design of MLLMs for video understanding. 
Cybernetics provides a rich theoretical framework centered on feedback loops, adaptive control, and goal-oriented behavior, enabling stable and effective operation in complex, dynamic settings. 
We hypothesize that by redesigning video MLLMs as cybernetic systems, capable of self-monitoring, self-correction, and adaptive resource allocation during inference, we can significantly enhance their performance.

Implementing these principles, we introduce \textbf{\name{}}, a framework structured as a dynamic feedback loop to create more robust and accurate MLLMs for video understanding. 
The core of this cybernetic system consists of three components: the MLLM Inference System, the Sensor and the Controller. 
Specifically, we generate responses by applying various scaling strategies within the MLLM inference system. These responses may come directly from the base MLLM model or be enhanced through techniques such as chain-of-thought prompting or the incorporation of key frames.
The sensor monitors the inference processes and collect intermediate interpretations, such as the attention drift among different outputs and the prediction results, as evidence for the re-analysis of the controller. 
Given the evidence, the controller calculates the confidence score and determines whether to terminate the loop or to forcibly trigger a self-correction process to avoid unreliable responses during inference. If the termination condition is not met, the controller takes action by injecting the generated feedback into the MLLM’s input for the next round of inference, closing the entire cybernetic loop.
We demonstrate the efficacy of the proposed cybernetically-inspired mechanisms on challenging video understanding benchmarks. 
Our experiments demonstrate that our method can remarkably improve the performance of a relatively small model by a large margin.
As shown in Figure~\ref{fig:teaser}, it improves the accuracy of Qwen2.5-VL-7B by 8.3\% and InternVL3-8B by 5.5\%, allowing both models to surpass GPT-4o on the VideoMMMU benchmark. 
Furthermore, when applied to a larger model, Qwen2.5-VL-72B, our approach yields a 10.0\% improvement over the baseline, achieving performance comparable to human experts. 
In addition to excelling on knowledge-centric benchmarks that benefit from logical reasoning, our method can also generalize to broader video domains: it delivers consistent gains on general-purpose benchmarks such as VideoMME and WorldSense, with 1.1\% improvements on both.
Extensive ablation studies show the effectiveness of each component in our \name{} framework.

Our \textbf{contributions} can be summarized as follows:

    \noindent 1. We propose \textbf{\name}, a test-time adaptive scaling framework based on cybernetic feedback control that enhances frozen MLLMs without training or extra components (vision expert models).
    
    \noindent 2. We introduce an \textbf{attention-based} monitoring mechanism and an adaptive scoring controller that jointly govern strategy selection during inference.
    
    \noindent 3. \name{} empowers small models to outperform proprietary systems like GPT-4o, and enables large open-source models to achieve state-of-the-art results on VideoMMMU. Extensive experiments demonstrate the effectiveness and the generalization capability of our approach.

\section{Related Work}
\label{sec:related_work}

\noindent
\textbf{Multi-modal Large Language Models in Video.}
Multi-modal large language models (MLLMs) have recently achieved significant success~\cite{qwen2.5vl,zhu2025internvl3,shen2025long,ma2025mllm, luo2024feast,wu2024controlmllm,rang2025eve,jie2024memory}, with the video domain receiving growing attention.
In recent years, numerous video-specific multimodal large language models (MLLMs)~\cite{llava-video, damonlpsg2025videollama3,damonlpsg2024videollama2,videochat2,videoxl,videollama,longva,wang2025internvideo2,han2024free} have been developed to tackle challenges inherent in video understanding, such as temporal reasoning and cross-frame alignment. 
Meanwhile, several open-sourced foundation models such as the Qwen-VL series~\cite{bai2023qwen,wang2024qwen2,qwen2.5vl}, InternVL series~\cite{chen2024internvl,internvl-1.5,internvl2.5,zhu2025internvl3} aim to unify image and video processing within a single framework. 
While these architectures demonstrate impressive perception capabilities, most existing MLLMs struggle with complex reasoning over video content.
To address this gap, recent efforts~\cite{guo2025deepseek,muennighoff2025s1,PPO,DPO,shao2024deepseekmath} introduce reinforcement learning and test-time scaling strategies to enhance language model reasoning, and have been extended to video understanding through models such as Video-R1~\cite{video-r1}, VideoChat-R1~\cite{li2025videochat}, and TinyLLaVA-Video-R1~\cite{zhang2025tinyllava}.
In parallel, video chain-of-thought (CoT) prompting methods such as Video-of-Thought~\cite{video-of-thought}, Chain-of-Shot~\cite{hu2025cos}, Logic-in-Frames~\cite{guo2025logic}, R3CoT~\cite{r3cot} decompose complex video reasoning tasks into manageable sub-problems, addressing them step-by-step from low-level perception to high-level cognition.
However, most existing approaches require supervised post-training or auxiliary models. 
In contrast, our work explores a training-free, single-model strategy that performs strongly on knowledge-intensive video tasks, suggesting that simple, modular inference techniques can yield robust multi-modal reasoning.

\noindent
\textbf{Test Time Scaling.} This direction~\cite{snell2024scaling,muennighoff2025s1,liu2025can,wei2022chain,brown2024large,levi2024simple, bi2024forest} is a promising strategy for improving LLM performance by allocating more compute during inference rather than increasing model size.
TTS methods generally fall into two categories: \textbf{Sequential Scaling}, which prolongs the reasoning process (e.g. chain of thought~\cite{wei2022chain}, reflection~\cite{muennighoff2025s1}); and \textbf{Parallel Scaling}, which explores multiple reasoning paths and selects the best (e.g. best-of-n~\cite{brown2024large,levi2024simple}). Parallel methods are often combined with sequential strategies to form complex search procedures, such as tree search~\cite{gandhi2024stream,liu2025can, bi2024forest}, with majority voting~\cite{wangself}, output reward models (ORMs)~\cite{xin2024deepseek,ankner2024critique} and process reward models (PRMs)~\cite{uesato2022solving,lightman2023let} often used to verify reasoning steps.
Recent works have instantiated these strategies in various ways: s1~\cite{muennighoff2025s1} adopts sequential scaling via a lightweight budget-forcing mechanism to control inference depth adaptively, while \cite{liu2025can} formulates compute-optimal policies for test-time search by balancing performance and computational cost.
While effective in textual tasks, TTS remains underexplored in video understanding. 
Our findings suggest that directly applying existing techniques often yields limited gains, highlighting the need for modality-aware scaling strategies.

\noindent\textbf{Cybernetics in Machine Learning and AI Systems.}
Cybernetics, first formalized by Wiener~\cite{wiener1948cybernetics}, provides a theoretical framework for self-regulating systems composed of three core components: a sensor for observing system states, a controller for decision-making, and a plant or system being controlled~\cite{wiener1948cybernetics, mcculloch1943logical, gage2007boat}. While these principles influenced early AI research~\cite{ashby1956introduction, mcculloch1943logical, foerster1952cybernetics}, their integration into modern deep learning remains limited~\cite{cariani2010importance}. Some recent works have introduced feedback mechanisms into neural architectures, such as CNN-F~\cite{huang2020neural}, which adds recurrent generative feedback to improve adversarial robustness, and AdaRefiner~\cite{zhang2023adarefiner}, which establishes a closed-loop interaction between language models and RL agents. However, these approaches often require architectural modifications or specialized training. In contrast, our approach utilizes the MLLM Inference System, Sensor, and Controller to apply cybernetic principles. By modeling video understanding with these components, we significantly improve performance without the need for additional training.
\section{Method}
\label{sec:method}

\subsection{Cybernetic View for Video Test-Time Scaling}
\label{sec:cyber_view}

Test-time scaling for multimodal large language models (MLLMs), particularly in video understanding tasks, presents significant challenges. Unlike text-only reasoning, where techniques such as chain-of-thought prompting can be applied with moderate success, MLLMs face greater complexity due to the temporal, visual, and semantic richness of video data. 
Existing approaches~\cite{qwen2.5vl, zhu2025internvl3, li2024llavaonevision, llava-video, video-of-thought, hu2025cos}
often apply static inference strategies that do not adapt to input difficulty, uncertainty, or reasoning failure, leading to inefficiencies and suboptimal performance. To overcome these limitations, we propose a cybernetic framework that transforms test-time inference into a feedback-driven, adaptive process inspired by the principles of control and regulation in cybernetics. We conceptualize video understanding during inference as a closed-loop control system consisting of three interdependent components:

% by mjh
% MLLM Inference System（被控系统）：MLLM进行视频理解的过程
% Sensor: 收集MLLM的输出，并监控 LLM forward中间量，从中提取一些信号比如attention差异，
% Controller: 利用score forest对每个结果打分；比较得到的回复置信度和目标，决定进行输出还是自我纠错；如果进行自我纠错，要提供正确的feedback，比如引入关键帧，重新组织MLLM的输入(action)

\textbf{MLLM Inference System:} This is the plant in the cybernetic loop, responsible for executing inference over multimodal input. These responses serve as raw material for further evaluation.

\textbf{Sensor:} The Sensor monitors the forward execution of the MLLM and extracts key signals such as predicted options and attention drift among different responses. These signals reflect the inference reliability, forming the basis for later decision-making.

\textbf{Controller:} The Controller is the central decision-making unit of the cybernetic system. It receives multiple signals from the Sensor, and evaluates the confidence of each candidate response using a rule-based scoring ensemble. Based on a thresholding policy, it decides whether to accept the output or trigger further inference with generated feedback.

% \textbf{Sensor:} The Sensor extracts key signals from the model's outputs to inform downstream control decisions. It identifies the predicted answer by parsing response texts and computes attention drift by comparing attention patterns across different inference strategies. These signals reflect the model’s attention stability, forming the basis for later decision-making.

% \textbf{Controller:} The Controller uses the signals provided by the Sensor and outputs from the Actuator to assess the uncertainty of the current predictions. 
% %
% It scores each candidate's response and aggregates these scores to evaluate overall confidence. 
% %
% Based on a thresholding policy, it decides whether to accept the output or trigger further inference, such as adding key frames to enhance visual clues.

% \textbf{Actuator:} The Actuator executes inference strategies based on the Controller's instructions. It interacts with the underlying model to produce candidate responses under different configurations, enabling the system to adaptively explore diverse interpretations or focus on different aspects of the video.

Specifically, given a video \( V \), a query \( q \) that includes the question and the video subtitles (if available), and a strategy set \(\Pi = \{ \pi_1, \dots, \pi_N \}\), the frozen model \(\mathcal{M}\) generates candidate responses $\{ r_i\}_{i=1}^N$, where each $r_i$ can be expressed as $r_i = \mathcal{M}_{\pi_i}(q, V).$
These responses are then passed back to the Sensor and Controller for evaluation, forming a closed feedback loop.
This iterative process allows the model to monitor its own interpretations, adjust inference paths in real time, and allocate computational resources more efficiently based on task relevance and video complexity. 
Unlike prior test time scaling approaches~\cite{muennighoff2025s1,wei2022chain} that apply fixed reasoning templates regardless of context, our method dynamically modulates processing depth and focus, enhancing both accuracy and robustness.

\begin{figure}[t]
    \centering
    \includegraphics[width=1.0\linewidth]{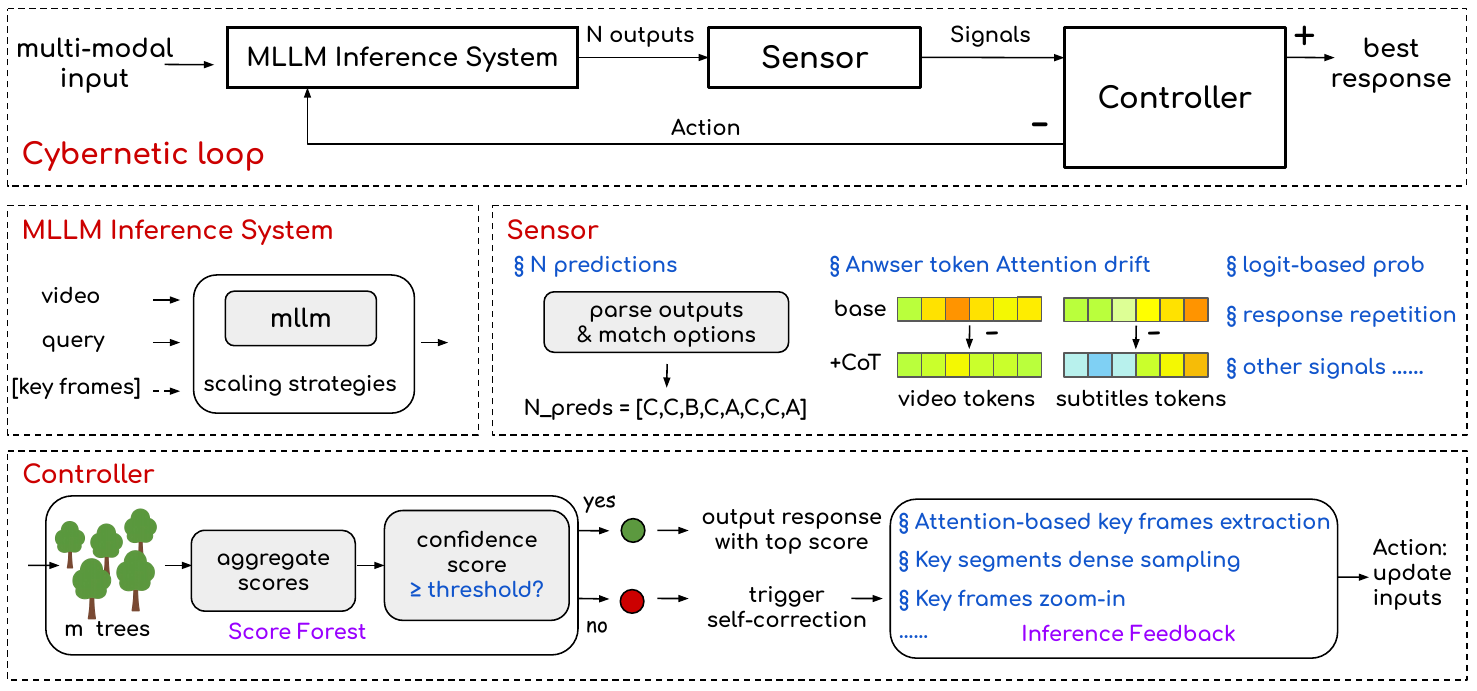}
    \caption{
    \small{
    \textbf{Overall framework of \name.} 
    \name{} models test-time video understanding as a closed-loop cybernetic process comprising three modules: \textit{MLLM Inference System}, \textit{Sensor}, and \textit{Controller}. The \textbf{MLLM Inference System} executes inference scaling strategies with a frozen MLLM on multi-modal input, generating N outputs. The \textbf{Sensor} monitors the forward process of the MLLM, extracting intermediate signals such as parsed prediction and attention drift. The \textbf{Controller} aggregates multiple signals to evaluate response reliability with \textbf{Score Forest} and triggers an self-correction actions when confidence falls below threshold through \textbf{Inference Feedback} module. The updated input is then used to re-invoke the MLLM. This feedback loop enables adaptive and robust test-time reasoning.
    }
    }
    \label{fig:framework}
\end{figure}

As illustrated in Figure~\ref{fig:framework}, the proposed framework \textbf{\name{}}
implements this \textbf{cybernetic loop} to adaptively scale inference at test time without any parameter updates or supervision. 
It empowers a frozen model to handle diverse video understanding tasks by actively managing its reasoning process in response to control signals.
The pseudo-code of \name{} is presented in the Appendix~\ref{sec:pseudo-code}.

\subsection{MLLM Inference System: Executing Test-time Scaling Strategies}
\label{sec:actuator}
The MLLM inference system, serving as the plant of the cybernetic loop, can execute diverse test-time scaling strategies over the multimodal input. 
We adopt the Best-of-N~(BoN) scheme that executes N forward passes in parallel to generate a set of candidate responses. Each inference path may vary in its configuration, including direct answer using the base model, chain-of-thought prompting to encourage reasoning, such as ``Thinking Process:'', or the incorporation of visually enhanced inputs such as injected key frames.
This design enables the system to adaptively combine structured reasoning and perceptual reinforcement based on task uncertainty. 
Compared to more complex search strategies, such as performing tree search, Best-of-N offers a simpler yet effective alternative.

\subsection{Sensor: Signal Extraction from MLLM Forward Processes}
\label{sec:sensor}
The Sensor monitors the forward execution of the MLLM and extracts informative signals that serve as the basis for confidence evaluation and feedback decisions.

One key signal is the predicted answer label $\{\hat{y}_n\}_{n=1}^N$, obtained by parsing each of the $N$ textual responses $\{r_n\}_{n=1}^N$, where $\hat{y}_n \in \mathcal{C}$, and $\mathcal{C}$ is the candidate set of choices in the predictions, e.g. A, B, C, D etc. The parsing relies on explicit pattern matching or approximate content alignment to handle free-form text.

Additionally, the Sensor evaluates the model’s perceptual behavior by quantifying attention drift. As the video and subtitles can be segmented according to the number of frames and timestamps, we compare the attention distribution over these segments across two settings: a base response and a chain-of-thought prompting variant.  
Specifically, the video is divided into \(K_1\) segments and the subtitles into \(K_2\) segments. For each attention head \(h \in \{1, \dots, H\}\), where \(H\) is the total number of attention heads, we define the attention scores from the answer token to the video and subtitle segments in the final layer as \(\mathbf{A}^{\text{video}}_h \in [0, 1]^{1 \times K_1}\) and \(\mathbf{A}^{\text{sub}}_h\in [0, 1]^{1 \times K_2}\). 
The attention drift signal $\Delta^{\text{video}} \in [-1, 1]^{1 \times K_1}$ for video part and $\Delta^{\text{sub}} \in [-1, 1]^{1 \times K_2}$ for subtitles part is defined as:
\begin{align}
\Delta^{\text{video}} = \frac{1}{H}\sum_{h=1}^{H} \left( \mathbf{A}^{\text{video}}_{h,\text{cot}} - \mathbf{A}^{\text{video}}_{h,\text{base}}\right),
\quad
\Delta^{\text{sub}} = \frac{1}{H}\sum_{h=1}^{H} \left( \mathbf{A}^{\text{sub}}_{h,\text{cot}} - \mathbf{A}^{\text{sub}}_{h,\text{base}} \right).
\end{align}
where the subscript ``base'' and ``cot'' refer to the base and chain-of-thought model responses, respectively. The subtitle part is not considered when the video has no audio. These attention differences reveal whether the model's focus has shifted away from the most relevant visual or subtitle segments during reasoning. A larger negative value of $\Delta$ suggests degraded perceptual grounding, prompting the controller to inject key frames that re-anchor the model’s attention to the critical visual evidence.

Beyond attention and answer prediction, the Sensor can also collect other forward-pass signals to characterize response quality. For example, the softmax confidence of the predicted token, logit stability across strategies, and repetition patterns in the response text. Together, these features offer a rich diagnostic view of the model’s current inference behavior and serve as input to the control mechanism that governs adaptive reasoning.

\subsection{Controller: Decision Making and Feedback Construction}
\label{sec:controller}

The Controller governs the adaptive reasoning process by making two key decisions: whether the current inference results are sufficiently reliable for output, and if not, how to generate actionable feedback to revise the model’s input for the next iteration. It comprises two coordinated modules: a \textbf{Score Forest} for response evaluation and an \textbf{Inference Feedback} module for corrective input construction.

\textbf{Score Forest: Confidence-Aware Evaluation and Thresholding.}  
Given $N$ candidate responses and multiple signals from the sensor, the Score Forest assigns each response a multi-dimensional score vector $\mathbf{s}_n = (s_{n,1}, \dots, s_{n,m}) \in [0,1]^{m}$, capturing semantic, probabilistic, and attention-related qualities, where $m$ is the number of trees in the forest and each tree maps the extracted signals to a score within the range of 0 to 1 via a distinct mechanism. In our implementation, the scoring mechanisms include, for instance, the softmax confidence of the predicted answer token, an indicator of logit stability across reasoning strategies, a binary repetition penalty reflecting output redundancy, a visual attention retention score quantifying how well visual grounding is preserved, and a normalized rank score that captures the relative quality of each candidate among all $N$ responses.
For each response \( n \in \{1, \dots, N\} \), we compute $S_n \in [0,1]$ as an aggregation of individual scores \( s_{n,i} \), weighted by hyper-parameters \( \{\beta_i\}_{i=1}^{m} \):
\begin{align}
S_n = \sum_{i=1}^{m} \beta_i s_{n,i}, \quad \sum_{i=1}^{m} \beta_i = 1.
\end{align}

Furthermore, we can calculate the top-scoring option based on these scores. The score of the best option, $TopScore \in [0, N]$, can be calculated by $ TopScore = \max_{c \in \mathcal{C}} \sum_{n : \hat{y}_n = c} S_n$.
If the top score satisfies the confidence threshold, i.e., $ TopScore \geq \tau \cdot N$, where the threshold $\tau \in [0,1]$, the controller selects the top-score answer as the final output. Otherwise, the low confidence score indicates unreliable initial reasoning, and the controller triggers the Inference Feedback module to initiate a corrective update.
Note that the widely used majority voting policy can be viewed as a special case of the Score Forest, where the score $S_n = 1$ for each response $n$ and the threshold $\tau=0$.

\textbf{Inference Feedback: Visual Correction for Self-Revision.}  When confidence is insufficient, the Controller invokes the Inference Feedback module to construct enhanced input that guides the next round of reasoning. This module identifies the top-k visual and subtitle segments that exhibit the greatest decrease in attention. Specifically, we define: $\mathcal{I}_{\text{video}} \subseteq \{1, \dots, K_1\}$ and $\mathcal{I}_{\text{sub}} \subseteq \{1, \dots, K_2\}$, 
where each set contains the indices of the top-k segments with the largest attention decrease:
\begin{align}
\mathcal{I}_{\text{video}} = \operatorname{TopK-Indices}(-\Delta_j^{\text{video}}), \quad
\mathcal{I}_{\text{sub}} =  \operatorname{TopK-Indices}(-\Delta_j^{\text{sub}}).
\end{align}

For $\mathcal{I}_{\text{video}}$, we can directly extract the corresponding frames through the indices. For $\mathcal{I}_{\text{sub}}$, we trace their timestamps to locate the aligned frames. The union of these yields the final set of key frames. 
To restore the model’s degraded attention due to reasoning steps, the identified key frames can be seamlessly re-integrated into the original input sequence.
To further refine the model’s focus, we also support more visual content enhancement methods. Temporally, we perform dense sampling around selected key frames while sparsely sampling elsewhere. Spatially, we compute region-question relevance and apply zoom-in to emphasize evidence-rich regions.
These enhanced inputs are then sent back to the MLLM inference system, enabling the model to refocus on critical evidence.

\section{Experiments}
\label{sec:exp}

\noindent
\textbf{Benchmarks.}
We conduct experiments on various video understanding benchmarks, grouped into two categories: knowledge-centric and general-purpose videos.
VideoMMMU~\cite{videommmu} represents the knowledge-centric category, consisting of 300 expert-level educational videos and 900 questions across six academic domains. 
It is well-suited for assessing models' reasoning and knowledge application. 
For general-purpose videos, we use VideoMME~\cite{videomme} and WorldSense~\cite{hong2025worldsense}. 
VideoMME includes 900 videos from six domains and 2,700 multiple-choice questions, while WorldSense contains 1,662 audio-visual videos and 3,172 questions spanning 26 task types, emphasizing multimodal integration. 
%
% Lastly, CGBench~\cite{chen2024cg} focuses on long-form video understanding, with 1,219 videos averaging over 27 minutes in length. 
%
Together, these benchmarks enable a comprehensive evaluation of model performance across diverse content types, temporal scales, and reasoning demands. 

\noindent
\textbf{Evaluation Metrics.}
We use accuracy as the evaluation metric for all benchmarks. For VideoMMMU, which includes both multiple-choice and open-ended questions, we follow its official protocol. 
For VideoMME, WorldSense, which contains only multiple-choice questions, accuracy is computed as the proportion of correct predictions.

\begin{table}[t]
  \centering
  \setlength{\tabcolsep}{5pt}
  \caption{\small{\textbf{Performance on the VideoMMMU benchmark (accuracy \%).} The results are grouped by evaluation track (Perception, Comprehension, Adaptation) and academic discipline (Art, Business, Science, Medicine, Humanities, and Engineering). Yellow rows indicate open-source MLLMs, while blue rows indicate proprietary models. Models marked with \textit{w/sub} indicate our baselines with subtitle input.}}
  \label{tab:videommmu-main}
  \scalebox{0.82}{
  \begin{tabular}{lcccccccccc}
    \toprule[0.15em]
    \textbf{Model} & \textbf{Overall} & \multicolumn{3}{c}{\textbf{Results by Track}} & \multicolumn{6}{c}{\textbf{Results by Discipline}} \\
    \cmidrule(lr){3-5} \cmidrule(lr){6-11}
     &  & \textbf{Percep.} & \textbf{Compr.} & \textbf{Adapt.} & \textbf{Art.} & \textbf{Biz.} & \textbf{Sci.} & \textbf{Med.} & \textbf{Hum.} & \textbf{Eng.} \\
    \midrule
    Human Expert & 74.4 & 84.3 & 78.7 & 60.3 & 81.0 & 78.8 & 74.2 & 70.5 & 84.8 & 69.9 \\
    \midrule
    \rowcolor{yellow!10} LLaVA-OneVision-7B~\cite{li2024llavaonevision} & 33.9 & 40.0 & 31.0 & 30.7 & 49.2 & 29.6 & 34.9 & 31.8 & 46.7 & 29.2 \\
    \rowcolor{yellow!10} VILA1.5-40B~\cite{lin2024vila}  & 34.0 & 38.7 & 30.7 & 32.7 & 57.1 & 27.3 & 23.5 & 38.0 & 41.9 & 32.5 \\
    \rowcolor{yellow!10} LLaVA-Video-7B~\cite{llava-video}  & 36.1 & 41.7 & 33.3 & 33.3 & 65.1 & 34.1 & 32.6 & 42.6 & 45.7 & 27.4 \\
    \rowcolor{yellow!10} InternVL2-8B~\cite{internvl-1.5} & 37.4 & 47.3 & 33.3 & 31.7 & 55.6 & 34.1 & 30.3 & 34.1 & 41.9 & 38.1 \\
    \rowcolor{yellow!10} LLaVA-OneVision-72B~\cite{li2024llavaonevision} & 48.3 & 59.7 & 42.3 & 43.0 & 61.9 & 46.2 & 40.2 & 54.3 & 60.0 & 44.0 \\
    \rowcolor{yellow!10} LLaVA-Video-72B~\cite{llava-video} & 49.7 & 59.7 & 46.0 & 43.3 & 69.8 & 44.7 & 41.7 & 58.9 & 57.1 & 45.1 \\
    \rowcolor{blue!10} Gemini 1.5 Flash~\cite{team2024gemini}  & 49.8 & 57.3 & 49.0 & 43.0 & 63.5 & 53.0 & 43.2 & 49.6 & 59.1 & 45.7 \\
    \rowcolor{yellow!10} Aria~\cite{li2024aria} & 50.8 & 65.7 & 46.7 & 40.0 & 71.4 & 47.7 & 44.7 & 58.9 & 62.9 & 43.7 \\
    \rowcolor{blue!10} Gemini 1.5 Pro~\cite{team2024gemini}  & 53.9 & 59.0 & 53.3 & 49.3 & 57.1 & 59.1 & 49.1 & 57.4 & 58.1 & 50.3 \\
    \rowcolor{yellow!10} Qwen2.5-VL-7B~\cite{qwen2.5vl} (w/ sub) & 55.0  & 72.7 & 53.7 & 38.7 & 73.0 & 56.1 & 46.2 & 58.1 & 73.3 & 47.8 \\
    \rowcolor{yellow!10} InternVL3-8B~\cite{zhu2025internvl3} (w/ sub) & 57.4 & 77.0 & 49.7 & 45.7 & 61.9 & 59.1 & 53.0 & 60.5 & 74.3 & 51.3 \\
    \rowcolor{blue!10} GPT-4o~\cite{openai2024gpt4o} & 61.2 & 66.0 & 62.0 & 55.7 & 69.5 & 66.9 & 51.6 & 64.8 & 69.5 & 57.1 \\
    \rowcolor{yellow!10} Qwen2.5-VL-72B~\cite{qwen2.5vl} (w/ sub) & 64.3 & 84.7 & 63.0 & 45.3 & 79.4 & 66.7 & 62.9 & 68.2 & 81.9 & 54.3 \\ 
    \rowcolor{blue!10} Claude 3.5 Sonnet~\cite{sonnet} & 65.8 & 72.0 & 69.7 & 55.7 & 66.7 & 75.0 & 56.1 & 58.1 & 75.2 & 66.1 \\
    \midrule
    % \rowcolor{green!10} 
    InternVL3-8B~\cite{zhu2025internvl3} (+Ours) & 62.9~(+5.5) & 77.3 & 60.3 & 51.0 & 65.1 & 67.4 & 62.1 & 62.0 & 80.0 & 56.0 \\
    % \rowcolor{green!10} 
    Qwen2.5-VL-7B~\cite{qwen2.5vl} (+Ours) & 63.3~(+8.3) & 78.0 & 62.0 & 50.0 & 76.2 & 65.9 & 54.5 & 64.3 & 75.2 & 59.3 \\
    % \rowcolor{green!10} 
    Qwen2.5-VL-72B~\cite{qwen2.5vl} (+Ours) & \textbf{74.3 (+10.0)} & \textbf{85.7} & \textbf{76.3} & \textbf{61.0} & \textbf{82.5} & \textbf{78.0} & \textbf{68.2} & \textbf{78.3} & \textbf{83.8} & \textbf{69.3} \\
    \bottomrule[0.15em]
  \end{tabular}
  }
\end{table}

\noindent
\textbf{Implementation Details.}
%\lxt{Too long and make it shorter. Make the more details in the appendix.} 
We adopt Qwen2.5-VL~\cite{qwen2.5vl} and InternVL3~\cite{zhu2025internvl3} as the base models for all evaluations, using 64 uniformly sampled frames per video for Qwen2.5-VL and 32 frames for InternVL3. Subtitles, when needed, are extracted using Faster-Whisper Large-v3.\footnote{\url{https://github.com/SYSTRAN/faster-whisper}}. We adopt two-round Best-of-N scheme across all benchmarks. In the Controller, response confidence is estimated via the Score Forest, and directly key frame injection is employed as the feedback mechanism. On the VideoMMMU dataset, we set $N=8$ and $\tau=0.3$ in the first round, using one base strategy and seven chain-of-thought (CoT) variants. In the second round, we set $N=1$ and $\tau=0$. All $\beta_i = \frac{1}{m}$ with $m=5$ in the experiments. Additional implementation details are provided in the Appendix~\ref{sec:more_implement}.
%

\begin{comment}
\begin{table}[t]
\centering
\caption{\small{Performance on general-purpose video understanding benchmarks (accuracy \%).\textit{w/ sub} means adding subtitles in the prompt.}}
\label{tab:other-benchmark-results}
\scalebox{0.85}{
\begin{tabular}{lccccc}
\toprule[0.15em]
\textbf{Model} & \multicolumn{4}{c}{\textbf{VideoMME (w/sub)}} & \multicolumn{1}{c}{\textbf{WorldSense (w/sub)}} \\
\cmidrule(lr){2-5} \cmidrule(lr){6-6}
 & Overall & Short & Medium & Long &  Overall\\
\midrule
Gemini 1.5 Pro~\cite{team2024gemini} & 81.3 & 84.5 & 81.0 & 77.4 & 39.3 \\
GPT-4o~\cite{openai2024gpt4o}  & 77.2 & 82.8 & 76.6 & 72.1 & 50.1 \\
\midrule
LLaVA-OneVision-7B~\cite{li2024llavaonevision} & 69.6 & 79.3 & 66.9 & 62.4 & 43.9 \\
Qwen2.5-VL-7B~\cite{qwen2.5vl} & 70.5 & 76.3 & 69.1 & 66.0 & 46.0 \\
Qwen2.5-VL-7B~\cite{qwen2.5vl} (+CoT) & 68.2~(-2.3) & 73.1 & 67.2 & 64.2 & 43.9~(-2.1) \\
Qwen2.5-VL-7B~\cite{qwen2.5vl} (Ours) & \textbf{71.6 (+1.1)} & 76.8 & 70.8 & 67.1 & \textbf{47.1 (+1.1)} \\
\bottomrule[0.15em]
\end{tabular}}
\end{table}
\end{comment}

\begin{table}[t]
\centering
\caption{\small{\textbf{Performance on general-purpose video understanding benchmarks (accuracy \%).} \textit{w/ sub} means adding subtitles. The two benchmarks are grouped based on video length and audio categories, respectively.}}
\label{tab:other-benchmark-results}
\scalebox{0.85}{
\begin{tabular}{lcccccccc}
\toprule[0.15em]
\textbf{Model} & \multicolumn{4}{c}{\textbf{VideoMME (w/sub)}} & \multicolumn{4}{c}{\textbf{WorldSense (w/sub)}} \\
\cmidrule(lr){2-5} \cmidrule(lr){6-9}
 & Overall & Short & Medium & Long & Overall & Speech & Event & Music \\
\midrule
Gemini 1.5 Pro~\cite{team2024gemini} & 81.3 & 84.5 & 81.0 & 77.4 & 39.3 & 39.6 & 38.9 & 39.2 \\
GPT-4o~\cite{openai2024gpt4o}  & 77.2 & 82.8 & 76.6 & 72.1 & 50.1 & 51.1 & 50.2 & 49.9 \\
\midrule
LLaVA-OneVision-7B~\cite{li2024llavaonevision} & 69.6 & 79.3 & 66.9 & 62.4 & 43.9 & 44.0 & 42.7 & 45.7 \\
Qwen2.5-VL-7B~\cite{qwen2.5vl} & 70.5 & 76.3 & 69.1 & 66.0 & 46.0 &  46.2 &  45.2 & 47.3\\
Qwen2.5-VL-7B~\cite{qwen2.5vl} (+CoT) & 68.2~(-2.3) & 73.1 & 67.2 & 64.2 & 43.9~(-2.1) & 43.9 & 43.5 & 44.6 \\
Qwen2.5-VL-7B~\cite{qwen2.5vl} (+Ours) & \textbf{71.6 (+1.1)} & 76.8 & 70.8 & 67.1 & \textbf{47.1 (+1.1)} & 47.4 & 46.6 & 48.0 \\
\bottomrule[0.15em]
\end{tabular}}
\end{table}

\subsection{Main Results}
\label{sec:exp_main_results}
% Logic / reasoning video : videommmu
% Short / Common video : videomme; worldsense
% Long video: cgbench

\noindent
\textbf{Knowledge-Centric Video Understanding.}
Table~\ref{tab:videommmu-main} shows results on the VideoMMMU benchmark. 
We compare the proposed method against two major categories of models: (1) open-source MLLMs, (2) proprietary models such as GPT-4o and Claude 3.5 Sonnet. Human expert performance is also provided for reference.
Our results show that \textbf{\name{} consistently enhances model performance} across a range of model scales. 
When applied to Qwen2.5-VL-7B, it achieves a notable +8.3\% improvement over the base model, surpassing GPT-4o by 2.1\% and approaching Claude 3.5 Sonnet. On InternVL3, it brings a +5.5\% gain, outperforming GPT-4o by 1.7\%. For Qwen2.5-VL-72B, \name{} further boosts performance by +10.0\%, exceeding Claude 3.5 Sonnet by 8.5\% and reaching accuracy on par with human experts. 
These results show that even smaller open-source models can outperform proprietary LLMs through effective cybernetic inference-time scaling.

In addition to the accuracy, we observe that \name{} is especially effective on \textbf{comprehension} and \textbf{application} tracks, where reasoning and knowledge transfer are essential. By discipline, the most significant gains occur in \textbf{business}, \textbf{science}, \textbf{medicine} and \textbf{engineering}.
These often require symbolic manipulation and mathematical deduction, indicating that our method is crucial for reasoning and knowledge-intensive video tasks.

\noindent
\textbf{General-Purpose Video Understanding.} Table~\ref{tab:other-benchmark-results} presents results on VideoMME and WorldSense, two benchmarks encompassing a wide range of everyday video types. 
As observed in this table, prompting the model to think before answering leads to degraded performance across both datasets. 
This highlights a key limitation of current multimodal models: due to insufficient cross-modal alignment, these models struggle to integrate visual observations, audio transcripts, and question semantics into coherent reasoning trajectories. 
In contrast to simple chain-of-thought prompting, \name{} consistently improves performance over the base model on general-purpose video benchmarks: +1.1\% on VideoMME and +1.1\% on WorldSense. 
While the absolute gains are modest, they highlight the value of content-aware test-time control that enhances model focus when initial reasoning is insufficient. 
As a result, \name{} demonstrates strong generalization across diverse video domains and offers a stable enhancement for test time scaling.

\subsection{Ablation Study on MLLM Inference System and Sensor.}
\label{sec:ablation-mllm-sensor}

We conduct a series of ablation experiments on the VideoMMMU benchmark to investigate the contribution of each component in the \name{} framework. All experiments are conducted under the zero-shot setting using Qwen2.5-VL-7B.

\noindent
\textbf{Performance boost via multiple inference strategies.}
As shown in Table~\ref{tab:MLLM-config}, directly answering with subtitles improves accuracy by 6.4\%, demonstrating the critical role of audio information in knowledge-centric video tasks. Adding the chain-of-thought prompt further boosts performance to 58.2\%, validating the effectiveness of textual reasoning. 
However, excessive reasoning may introduce distraction and attention drift. By directly incorporating attention-guided key frames augmentation based on our cybernetic system, performance further improves to 60.0\%. Note that here we adopt the \textbf{simplest form} of our framework: one base and one CoT response in the first round, followed by one response with key frames in the second round.

\noindent
\textbf{BoN outperforms complex search schemes.}
We further justify our choice of Best-of-N as the core inference framework. Beyond Best-of-N, PRM-guided tree search is a widely used scaling strategy that decomposes the reasoning process into multiple steps, where M candidates are selected from N at each step. As shown in Table~\ref{tab:MLLM-scaling-strategy}, tree-based search methods yield some improvement, but still fall short of BoN (N=8). We further investigate the impact of different values of N in the Appendix~\ref{sec:more_results}.

\noindent
\textbf{Sensor benefits from different attention sources.}
The Sensor module extracts intermediate signals such as attention drift. We compare two configurations under a fixed frame budget: one using only video part attention, and one also incorporating subtitle-guided attention based on timestamp alignment. As shown in Table~\ref{tab:sensor-attn}, subtitle-based drift offers a slight gain, indicating its complementary grounding value. However, for questions lacking clear temporal anchors, subtitle-aligned frames may introduce noise, suggesting the need for more refined filtering.

\begin{table}[t]
\centering
\caption{\small{\textbf{Ablation Study on MLLM Inference System and Sensor.} Accuracy (\%) on VideoMMMU under different inference configurations and search schemes for MLLM inference system, and different sources of attention drift for Sensor.}}
\label{tab:ablation-1}

\begin{tabular}{ccc}
\begin{minipage}{0.34\textwidth}
\begin{subtable}[t]{\linewidth}
\centering
\caption{\small{Impact of different scaling strategies. ``+'' denotes cumulative addition.}}
\label{tab:MLLM-config}
\scalebox{0.82}{
\begin{tabular}{l|l}
\toprule[0.15em]
\textbf{Strategy} & \textbf{Acc (\%)} \\
\midrule
Base                     & 48.6 \\
+Subtitles               & 55.0~(+6.4) \\
+CoT                     & 58.2~(+9.6) \\
+Key Frames              & 60.0~(+11.4) \\
\bottomrule[0.15em]
\end{tabular}
}
\end{subtable}
\end{minipage}
&
\begin{minipage}{0.26\textwidth}
\begin{subtable}[t]{\linewidth}
\centering
\caption{\small{Comparison of more search schemes. ``KF'' means key frames.}}
\label{tab:MLLM-scaling-strategy}
\scalebox{0.82}{
\begin{tabular}{l|c}
\toprule[0.15em]
\textbf{Scheme} & \textbf{Acc (\%)} \\
\midrule
Base~(+CoT\&KF)       & 60.0 \\
% +Depth                & 59.7 \\
+Best of N (Ours)     & 63.3 \\
+Tree Search          & 62.8 \\
% +Tree Search (PRM)    & 60.9 \\
\bottomrule[0.15em]
\end{tabular}
}
\end{subtable}
\end{minipage}
&
\begin{minipage}{0.30\textwidth}
\begin{subtable}[t]{\linewidth}
\centering
\caption{\small{Comparison of different sources of attention drift. ``+'' denotes cumulative addition.}}
\label{tab:sensor-attn}
\scalebox{0.82}{
\begin{tabular}{l|c}
\toprule[0.15em]
\textbf{Attn Source} & \textbf{Acc (\%)} \\
\midrule
Base (+CoT)        & 58.2 \\
+ Video-Part       & 59.9 \\
+ Subtitles-Part   & 60.0 \\
\bottomrule[0.15em]
\end{tabular}
% \begin{tabular}{c|c}
% \toprule[0.15em]
% \textbf{\#Paths N} & \textbf{Acc (\%)} \\
% \midrule
% % 1  & 55.0 \\
% 2  & 60.0 \\
% 4  & 60.9 \\
% 8  & 63.3 \\
% 16 & 62.7 \\
% 32 & 62.7 \\
% \bottomrule[0.15em]
% \end{tabular}
}
\end{subtable}
\end{minipage}
\end{tabular}
\end{table}

\subsection{Ablation Study on Controller.}
\label{sec:ablation-sensor-controller}

\noindent
\textbf{Score Forest outperforms majority voting.}
We evaluate the Controller's ability to make decisions based on uncertainty signals. Under the Best-of-N (N=8) setting, our score forest which aggregates multi-dimensional uncertainty outperforms simple majority voting (62.8\% vs. 61.9\% shown in Table~\ref{tab:controller-ablation}). With additional visual clues incorporated in next loop, our method further improves to 63.3\%. These results demonstrate that principled confidence modeling is more effective than uniform voting in test-time scaling, and confirm the critical role of the Controller in our cybernetic loop.

\noindent
\textbf{Different visual self-correction methods are effective.}
In addition to adding key frames directly, we further analyze the impact of more visual self-correction methods in the Controller. As shown in Table~\ref{tab:visual-control}, enhancing visual clues based on key frames with temporal dense sampling yields 60.3\%, while spatial zoom-in achieves the best performance at 60.7\%. These results validate the effectiveness of multi-dimensional visual scaling in improving model focus and answer accuracy. Due to the additional complexity of these methods, we use direct key frames injection in the main experiments.

\begin{table}[t]
\centering
\begin{minipage}[t]{0.64\textwidth}
\centering
\caption{\small{\textbf{Ablation Study on Sensor and Controller.} Accuracy (\%) on VideoMMMU under different scoring policies (Score Forest) and visual self-correction methods (Inference Feedback) in Controller.}}
\label{tab:ablation-controller}
\begin{subtable}[t]{0.48\linewidth}
\centering
\caption{\small{Scoring policies.}}
\label{tab:controller-ablation}
\scalebox{0.75}{
\begin{tabular}{l|c}
\toprule[0.15em]
\textbf{Scoring Method} & \textbf{Acc (\%)} \\
\midrule
Base (+CoT)                & 58.2 \\
Majority Voting            & 61.9 \\
Score Forest (One round)    & 62.8 \\
Score Forest (Ours)        & 63.3 \\
\bottomrule[0.15em]
\end{tabular}
}   
\end{subtable}
\hfill
\begin{subtable}[t]{0.48\linewidth}
\centering
\caption{\small{Visual self-correction methods.}}
\label{tab:visual-control}
\scalebox{0.75}{
\begin{tabular}{l|c}
\toprule[0.15em]
\textbf{Method} & \textbf{Acc (\%)} \\
\midrule
Base (+CoT)              & 58.2 \\
+ Key Frames             & 60.0 \\
+ Dense Sampling          & 60.3 \\
+ Spatial Zoom-in          & 60.7 \\
\bottomrule[0.15em]
\end{tabular}
}
\end{subtable}
\end{minipage}
\hfill
\begin{minipage}[t]{0.34\textwidth}
\centering
\vspace{0.2em}
\caption{\small{\textbf{Stability analysis under different disturbance levels.} ``+Ours'' refers to the simplest form mentioned in Section~\ref{sec:ablation-mllm-sensor}.}}
\label{tab:stability-analysis}
\scalebox{0.75}{
\begin{tabular}{l|c|c}
\toprule[0.15em]
\textbf{Setting} & \textbf{Base} & \textbf{+Ours} \\
\midrule
Uniform sampling       & 55.0 & 60.0 \\
Disturb rate = 0.2     & 55.0 & 60.4 \\
Disturb rate = 0.4     & 55.4 & 60.2 \\
Disturb rate = 0.6     & 52.0 & 60.1 \\
\bottomrule[0.15em]
\end{tabular}
}
\end{minipage}
\end{table}

\subsection{Stability Analysis}

To assess the robustness of our framework, we conduct a stability analysis inspired by control theory. We introduce temporal perturbations by replacing uniform frame sampling with a non-uniform variant, where each frame index is randomly shifted within a range determined by the disturb rate.
As shown in Table~\ref{tab:stability-analysis}, our method consistently outperforms the baseline across all disturbance levels, even as the baseline degrades. These results demonstrate that our method is stable and robust to non-uniform temporal distortions, confirming the strength of our cybernetic test-time scaling strategy in dynamically adapting to sampling perturbations while maintaining reliable performance.

\subsection{Visualization}

\begin{figure}[t]
    \centering
    \includegraphics[width=0.96\linewidth]{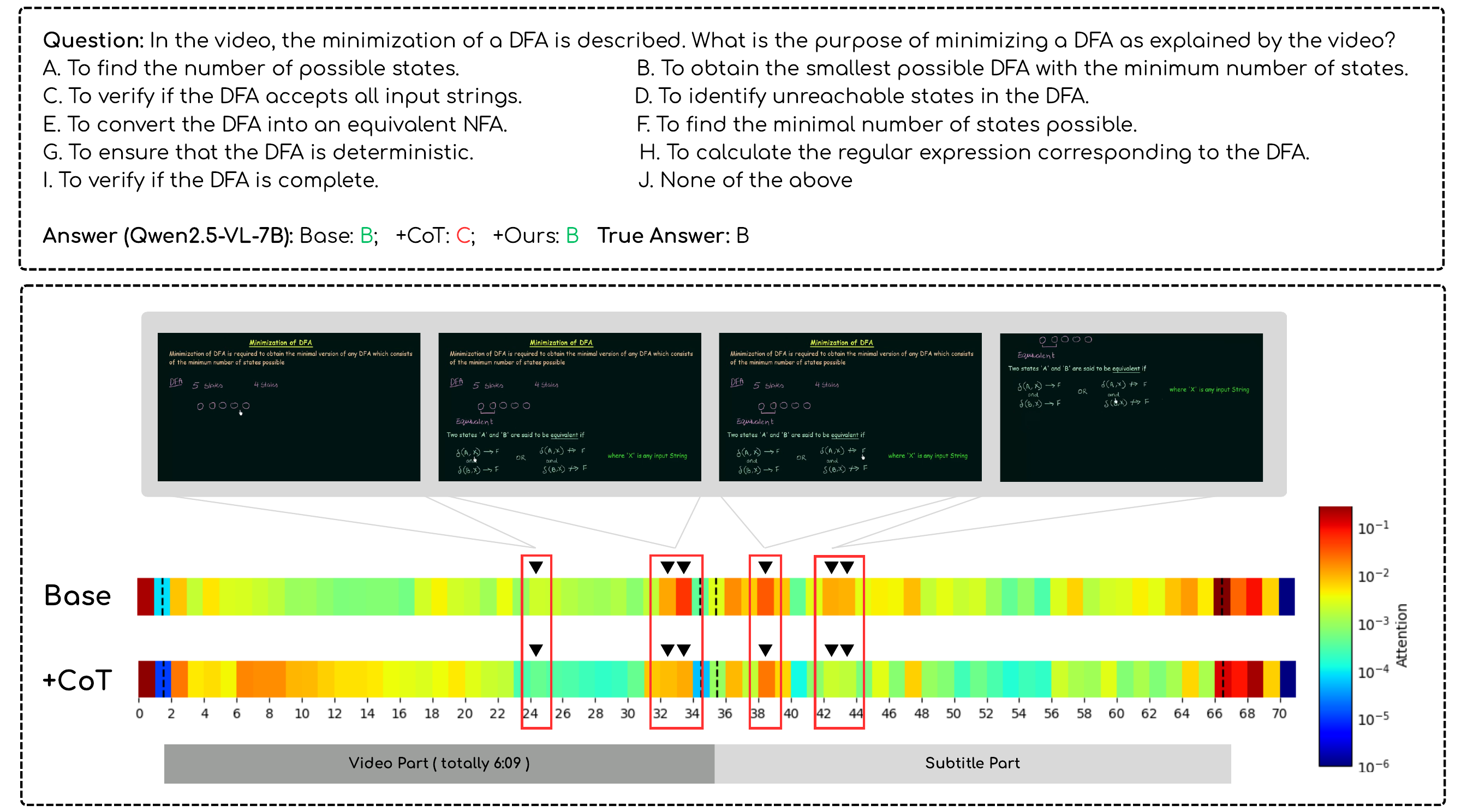}
    \caption{\small{Attention map visualizations on VideoMMMU. Red boxes highlight segments where attention significantly drops after applying ``CoT''. They may correspond to content that contains critical information. Under our control system, adding key frames after ``CoT'' helps rectify previously incorrect responses.}
}
    \label{fig:attn_map_good}
\end{figure}

Figure~\ref{fig:attn_map_good} illustrates the effectiveness of our control system in identifying forgotten yet critical visual information via attention difference after applying CoT. By reintegrating these cues, the system is able to correct CoT-induced errors, demonstrating the effectiveness of the self-correction mechanism. More visualizations and limitations of our work are provided in the Appendix~\ref{sec:more_visualization},\ref{sec:lim_future}.

\section{Conclusion}
\label{sec:conclusion}

We propose \name{}, a training-free, extra-model-free, test-time adaptive scaling framework designed to enhance video understanding performance in multimodal large language models (MLLMs). 
Inspired by cybernetic principles, \name{} integrates a closed-loop architecture with a MLLM Inference System-Controller-Actuator design to monitor attention shifts, evaluate prediction uncertainty, and dynamically execute self-correction strategies. 
Extensive experiments across diverse video benchmarks demonstrate the effectiveness of \name{}, achieving substantial gains on knowledge-centric and general-purpose video understanding tasks. 
Future work will explore more effective and efficient strategies to further improve complex multimodal reasoning.

\noindent
\textbf{Acknowledgement.} This work is supported by the National Key Research and Development Program of China (No. 2023YFC3807600).

\clearpage

\bibliographystyle{plainnat}
\bibliography{main}

\clearpage
\beginappendix
% \clearpage
% \setcounter{page}{1}
% \maketitlesupplementary

% \newpage
% \appendix

% \section{Appendix}
% \label{sec:appendix}

\section{Overview}

In this appendix, we first provide more implementation details of our method in \ref{sec:more_implement}. Then, we present additional experimental results and ablation studies on several component designs in \ref{sec:more_results}. Next, we include the PyTorch-style pseudo-code of our algorithm in \ref{sec:pseudo-code} to offer a clearer understanding of the overall inference process. After that, we present more visualization results in \ref{sec:more_visualization}. Finally, we discuss the limitations and future work in \ref{sec:lim_future}.

\section{More Implementation Details}
\label{sec:more_implement}

\noindent
\textbf{Inference details on each benchmark.} For all benchmarks, we adopt a two-round control loop with a fixed score aggregation weight $\beta_i = \frac{1}{m}$ and number of scoring trees $m = 5$. 
For \textbf{VideoMMMU}~\cite{videommmu}, we set $N = 8$ and $\tau = 0.3$ in the first round, including one base model response and seven chain-of-thought (CoT) variants. The base response uses deterministic decoding with temperature = 0.0, while CoT responses use sampling with temperature = 1.0, top-p = 0.5, and top-k = 5. In the second round, we set $N = 1$ and $\tau = 0.0$, applying a strategy that first generates a CoT reasoning trace, then augments it with attention-guided key frames before producing the final answer. During key frames selection, we use $\operatorname{TopK\text{-}Indices}$ with $K = 5$ for both video and subtitle segments. The total number of key frames is capped at 20 to control computational cost.
For \textbf{VideoMME}~\cite{videomme}, we adopt the simplest form of the cybernetic loop: in the first round, we perform one base and one CoT inference, using temperature = 0.0 and $\tau = 0.5$. The second round also performs a single pass, incorporating key frames augmentation to guide the final prediction.
For \textbf{WorldSense}~\cite{hong2025worldsense}, we set $N = 16$ and $\tau = 0.7$ in the first round, consisting of one base and fifteen CoT responses, all generated using temperature = 1.0. In the second round, we set $N = 8$, apply key frames augmentation, and rescore the candidate responses to obtain the final output.
For \textbf{MVBench}~\cite{li2024mvbench}, with results shown in Appendix~\ref{sec:more_results}, we set $N = 8$ in the first round with $\tau = 0.7$ and temperature = 1.0. The second round follows the same setting as that used for VideoMME.
All experiments are conducted on 8 GPUs, each equipped with 80 GB of memory.

\section{More Results and Ablation studies}
\label{sec:more_results}
\noindent
\textbf{Results on MVBench.}
We also evaluate our method on MVBench~\cite{li2024mvbench}, a challenging general-purpose video benchmark specifically designed to evaluate temporal comprehension in video understanding. As shown in Table~\ref{tab:mvbench-benchmark-results}, naive chain-of-thought prompting degrades Qwen2.5-VL-7B's performance by 4.6\%, dropping to 62.1\%. This highlights the challenges for multimodal large models to make effective reasoning on video understanding tasks, which is consistent with the results observed in other mentioned genral-purpose benchmarks. In contrast, the proposed approach \name{} yields improvements, boosting Qwen2.5-VL-7B's performance to 67.5\%. The results further demonstrate that our method effectively generalize across diverse video domains.

\begin{table}[h]
\centering
\caption{\small{\textbf{Performance on MVBench (accuracy \%).} }}
\label{tab:mvbench-benchmark-results}
\scalebox{0.99}{
\begin{tabular}{lc}
\toprule[0.15em]
\textbf{Model} & \textbf{Overall}  \\
\midrule
Gemini 1.5 Pro~\cite{team2024gemini} &  60.5 \\
GPT-4o~\cite{openai2024gpt4o} & 64.6 \\
\midrule
LLaVA-OneVision-7B~\cite{li2024llavaonevision} & 56.7  \\
Qwen2.5-VL-7B~\cite{qwen2.5vl} & 66.7 \\
Qwen2.5-VL-7B~\cite{qwen2.5vl} (+CoT) & 62.1 \\
Qwen2.5-VL-7B~\cite{qwen2.5vl} (Ours) & \textbf{67.5 (+0.8)} \\
\bottomrule[0.15em]
\end{tabular}}
\end{table}

\noindent
\textbf{Comparison of number of paths in BoN.}
We also study the effect of varying the number of inference paths N in the BoN setting. As shown in Figure~\ref{tab:ablation-bon}, performance steadily improves with larger N and saturates around N=8, which offers the best trade-off between diversity and computational cost.  

\noindent
\textbf{Comparison of attention extraction from different number of layers. }
Recent studies in model interpretability suggest that the final layers of large language models tend to capture more high-level semantic information that directly contributes to the model’s output decisions~\cite{ferrando2024know}. Motivated by this, we evaluate the effectiveness of extracting attention signals from different depths of the LLM component in Qwen2.5-VL-7B, which contains 28 transformer layers in total. Specifically, we experiment with extracting attention maps from the last 1, 4, and 7 layers, and report the results in Table~\ref{tab:attention-layers}.
Extracting attention solely from the final layer yields an accuracy of 60.0\%. Including the last 4 layers slightly reduces performance to 59.4\%, while using the last 7 layers gives a marginal improvement to 60.2\%. Overall, incorporating more layers introduces minor fluctuations, but yields consistent improvement over the baseline. For simplicity and computational efficiency, we finally extract attention only from the last layer in all experiments.

% % https://www.zhihu.com/question/11720422578
% % https://arxiv.org/pdf/2411.14257

% Table~\ref{tab:attention-layers} compares the model performance when attention is extracted from different numbers of final layers. Extracting attention solely from the last layer yields an accuracy of 60.0\%, while using the last 4 layers slightly decreases performance to 59.4\%. Utilizing the last 7 layers results in a marginal improvement to 60.2\%.

% These results suggest that using more layers for attention extraction does not necessarily improve performance and may introduce noise that weakens semantic focus. 

% These results suggest that increasing the number of layers used for attention extraction does not consistently enhance performance, and may introduce redundant or noisy information that dilutes the semantic focus of the model.

% In the context of our CyberV framework, where attention drift is a key signal used by the Sensor module to assess perceptual grounding, the reliability and specificity of attention features are critical. The findings indicate that selectively extracting attention from a smaller set of higher transformer layers may provide more effective signals for the downstream decision-making processes in adaptive inference systems.

\section{Pseudo-code of \name{}}
\label{sec:pseudo-code}

Figure~\ref{fig:code} illustrates the PyTorch-style pseudo-code of the \name{} architecture, which models video understanding as a closed-loop control process. It consists of three core modules: the MLLM Inference System executes various scaling strategies; the Sensor monitors intermediate outputs and extracts key signals such as predictions and attention drift; and the Controller evaluates response reliability and, if needed, constructs feedback (e.g., key frames augmentation) to trigger self-correction. These components interact iteratively to improve reasoning quality without additional training.

\begin{figure}[t]
\centering
\begin{tcolorbox}[
    title=Pseudo-code of the CyberV architecture.,
    colframe=gray!80!black,
    colback=white,
    fonttitle=\bfseries,
    sharp corners,
    boxrule=0.5pt,
    left=2pt,right=2pt,top=3pt,bottom=2pt,
    enhanced,
    width=\textwidth
]

\begin{multicols}{2}
\begin{lstlisting}[style=mystyle]
# Core module: Sensor
# signal extraction from MLLM forward processes
class Sensor:
    def monitor(outputs):
        signals = []
        signals.append(get_predicitons(outputs))
        signals.append(get_attn_drift(outputs))
        ...
        return signals
        
# Core module: Controller
# decision making and feedback construction
class Controller:
    def score_forest(out, attn_drift):
        # cal score for each response
        # aggregate scores for each option
        return score_list
        
    def inference_feedback(signals):
        # key frames extraction
        video_based_kfs, sub_based_kfs = ...
        kf =  video_based_kfs | sub_based_kfs
        # visual cues enhancement
        # Here is an example: directly augment key frames
        action = ("Add key frames.", kf)
        return action
        
    def decide(signals):
        scores = score_forest(signals)
        response = best_answer(scores)
        if is_confident(scores):
            return True, response, None
        else:
            action = inference_feedback(signals)
            return False, response, action 
\end{lstlisting}

\columnbreak
\begin{lstlisting}[style=mystyle]
# Core module: MLLM Inference System
# Executing Inference Strategies
class MLLMSystem:
    # an MLLM. For example, Qwen2.5-VL-7B
    self.model = ...
    def execute(inputs):
        outputs = []
        # Use BoN to get N responses
        for s in inputs['strategies']:
            out = self.model.forward(inputs, s)
            outputs.append(out)
        return outputs

# Cybernetic loop for one inference round
def run_one_loop(inputs):
    # Step 1: Run MLLM with multiple strategies
    out = MLLMSystem.execute(inputs)
    # Step 2: Monitor outputs to extract signals
    signals = Sensor.monitor(out)
    # Step 3: Decide when & how to trigger self-correction
    flag, response, action = Controller.decide(signals)
    return flag, response, action

# Closed-loop inference over multiple rounds
def cyber_v(inputs):
    max_rounds = ...
    round_now = 1
    while True:
        flag, response, action = run_one_loop(inputs)  
        if flag or round_now == max_rounds:
            return response  
        inputs = update(inputs, action)
        round_now += 1
\end{lstlisting}

\end{multicols}

\end{tcolorbox}
\vspace{-0.3em}
\caption{\small{Pseudo-code of the CyberV architecture. The MLLM Inference System, Sensor and Controller cooperate to form a closed-loop control cycle for test-time video understanding.}}
\label{fig:code}
\end{figure}

\begin{table}[t]
\centering
\begin{minipage}[t]{\textwidth}
\caption{\small{\textbf{More ablation studies.} Performance on VideoMMMU under different model settings.}}
\label{tab:ablation-study}
\begin{subtable}[t]{0.43\linewidth} 
\centering
\caption{Different number of paths (N) in BoN.}
\label{tab:ablation-bon}
\scalebox{0.95}{
\begin{tabular}{c|c}
\toprule[0.15em]
\textbf{\#Paths N} & \textbf{Accuracy (\%)} \\
\midrule
2          & 60.0  \\
4          & 60.9  \\
8          & 63.3  \\
16         & 62.7  \\
32         & 62.7  \\
\bottomrule[0.15em]
\end{tabular}
}
\end{subtable}
\hfill
\begin{subtable}[t]{0.55\linewidth}
\centering
\caption{Attention extraction from different layers. Here, we use Qwen2.5-VL-7B and compare the last 1st, 4th, and 7th layers.}
\label{tab:attention-layers}
\scalebox{0.95}{
\begin{tabular}{l|c}
\toprule[0.15em]
\textbf{Attention from} & \textbf{Accuracy (\%)} \\
\midrule
Base (No key frame)                & 58.2 \\
Last 1 layer                       & 60.0 \\
Last 4 layers                      & 59.4 \\
Last 7 layers                      & 60.2 \\
\bottomrule[0.15em]
\end{tabular}
}
\end{subtable}
\end{minipage}
\end{table}

\begin{figure}[t]
    \centering
    \includegraphics[width=1\linewidth]{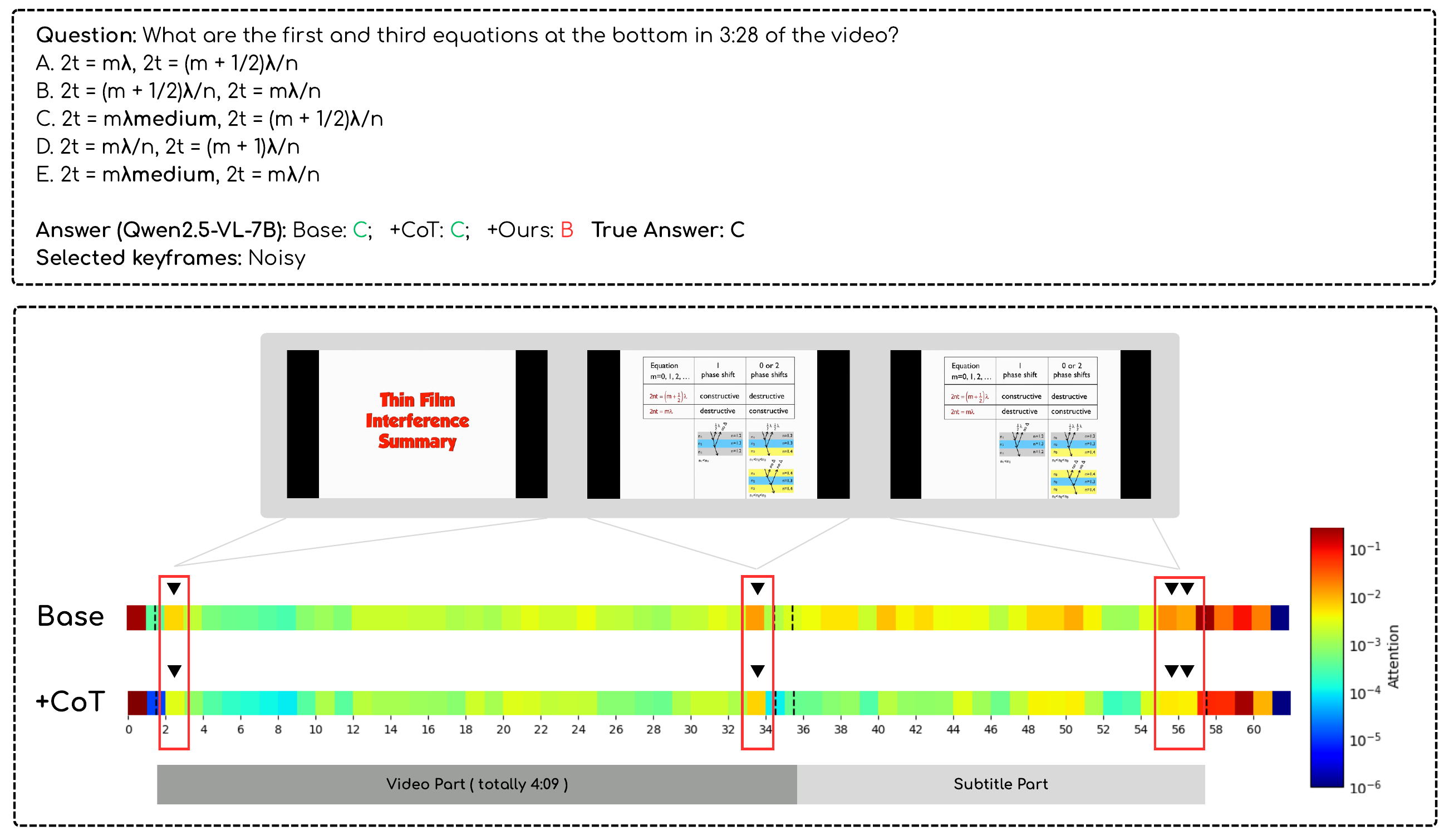}
    \caption{\small{Without confidence-based filtering in the Controller, high-confidence correct answers in the first round also need to trigger unnecessary key frames extraction, leading to errors in the second round due to noisy frames. In this case, although the question refers to 3:28 of a 4:09 video, the selected key frames focus on the beginning and end, resulting in an incorrect revision.}}
    \label{fig:attn_map_bad}
\end{figure}

\vspace{-2pt}
\section{More Visualization Results}
\vspace{-2pt}

\label{sec:more_visualization}

\noindent
\textbf{More attention map visualizations.} 
In the main paper, we show that incorporating attention-guided key frames into the second-round inference can effectively correct CoT-induced errors, particularly in cases where the base model initially produces the correct answer but CoT leads to an incorrect one. This demonstrates the utility of the cybernetic feedback loop in mitigating reasoning drift.

However, as illustrated in Figure~\ref{fig:attn_map_bad}, applying visual self-correction indiscriminately (without confidence-based filtering) can introduce new errors. In this example, both the base model and CoT initially provide the correct answer with high confidence, yet the second round reversed the decision due to the influence of noisy and irrelevant key frames. The contrast between this failure case and successful examples underscores the importance of incorporating confidence-aware control to selectively trigger feedback only when necessary, thereby enhancing overall robustness and preserving performance on easy cases.

\begin{figure}[t]
    \centering
    \includegraphics[width=0.9\linewidth]{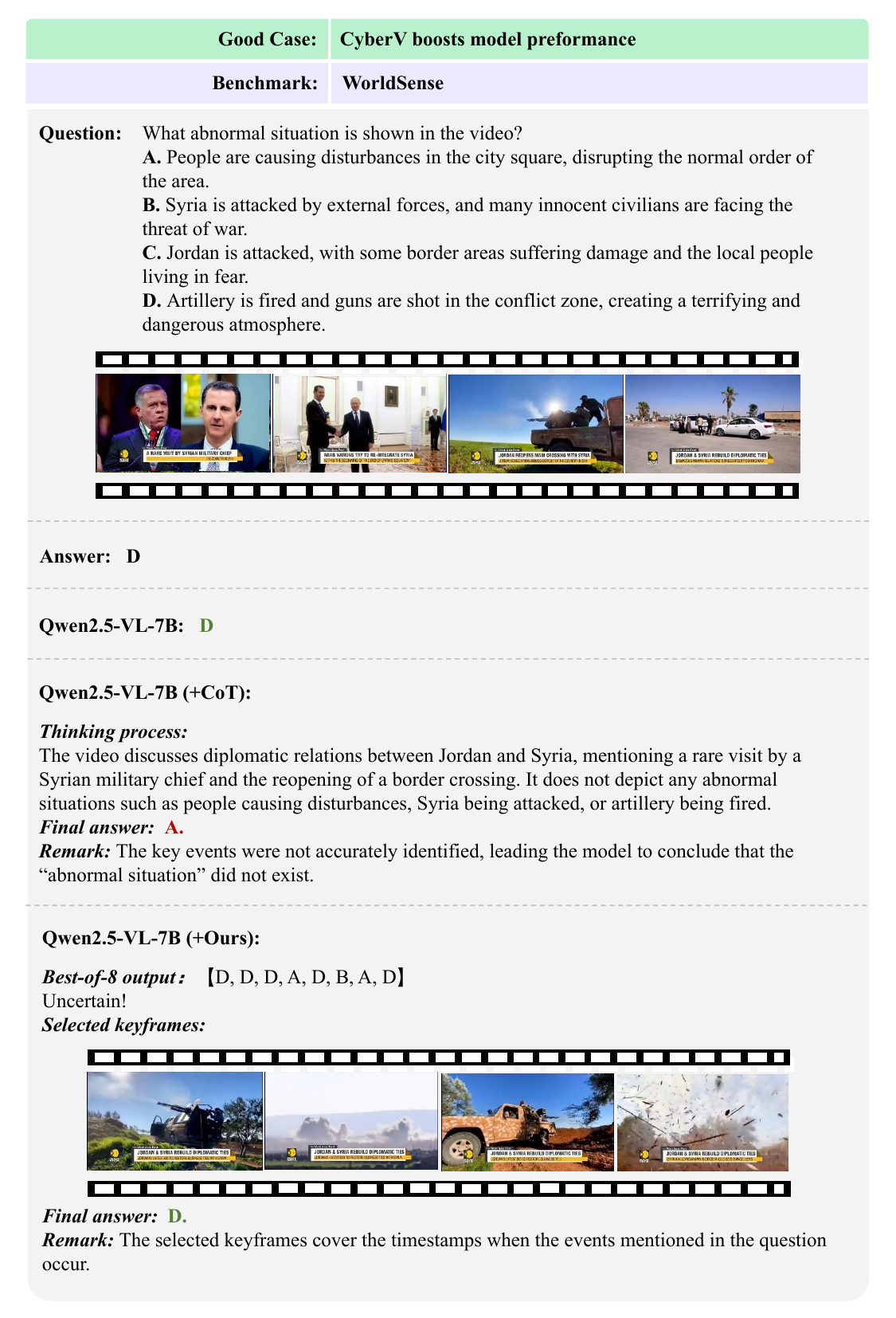}
    \caption{\small{CoT fails to answer but our method performs well.}}
    \label{fig:good_case_2}
\end{figure}

\begin{figure}[t]
    \centering
    \includegraphics[width=0.9\linewidth]{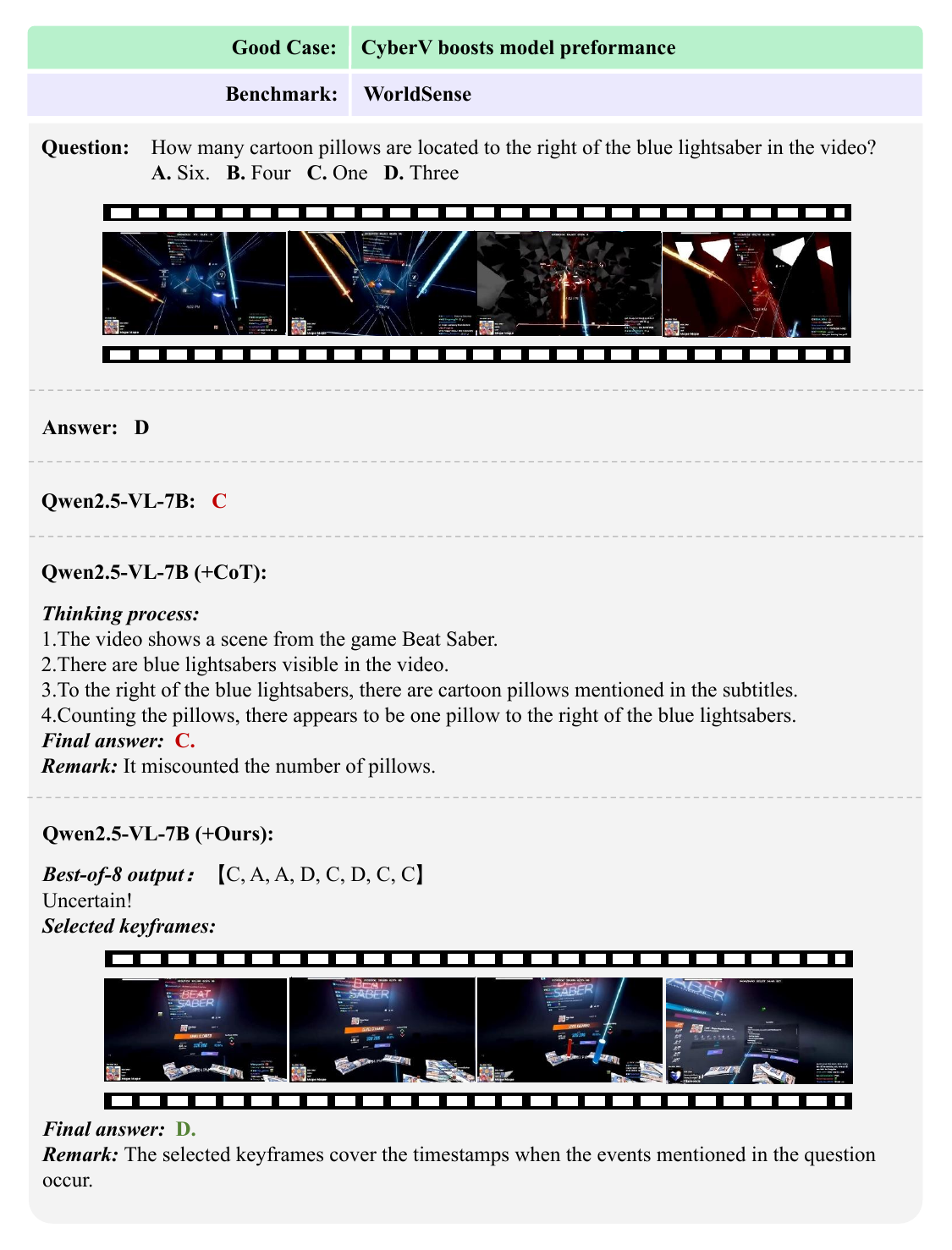}
    \caption{\small{Both the base model and CoT fail to answer but our method performs well.}}
    \label{fig:good_case_3}
\end{figure}

\begin{figure}[t]
    \centering
    \includegraphics[width=0.9\linewidth]{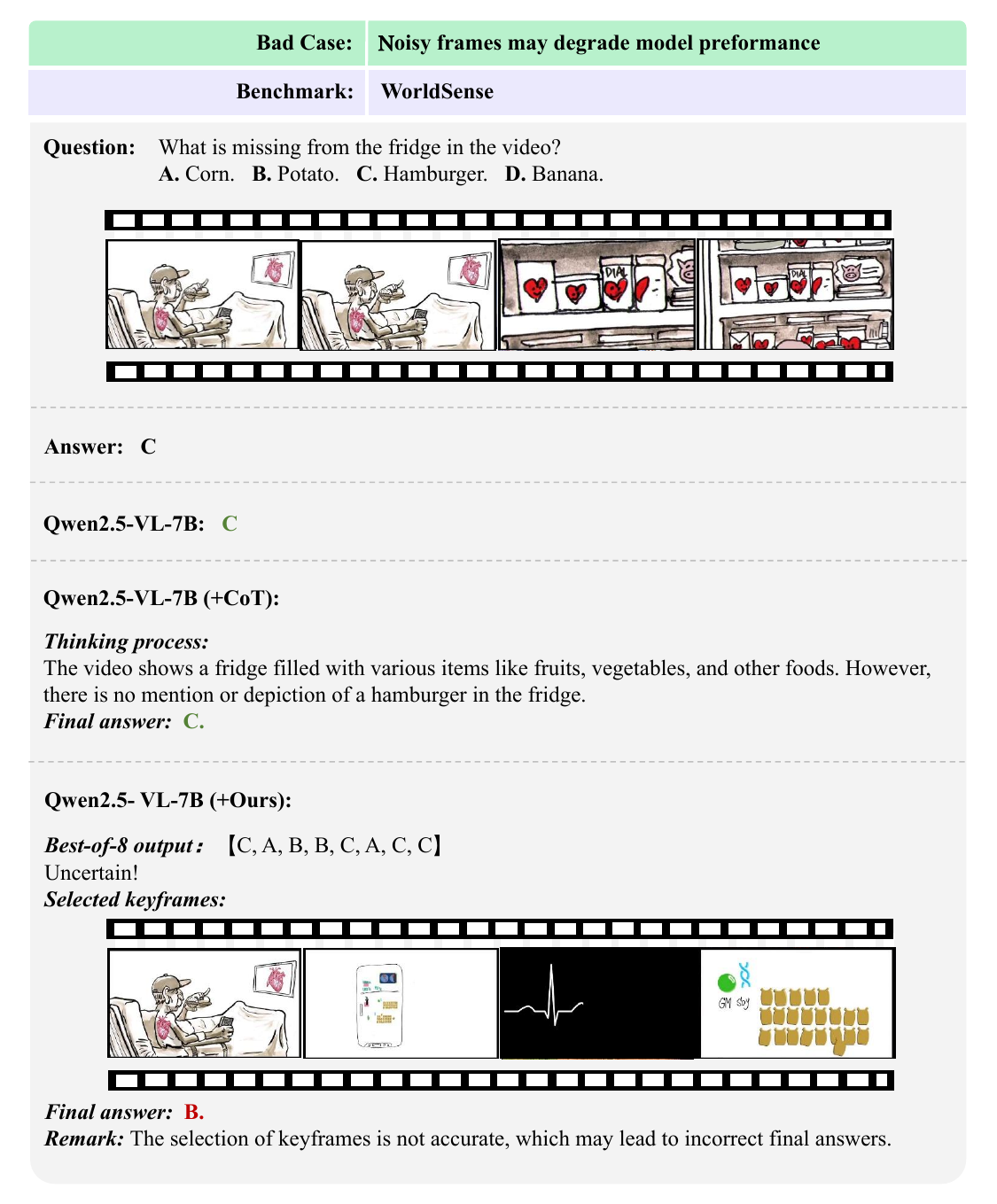}
    \caption{\small{Noisy frames may degrade model performance.}}
    \label{fig:bad_case_1}
\end{figure}

\noindent
\textbf{Case studies on WorldSense.}  
We conduct visualization studies on the WorldSense benchmark to qualitatively evaluate the effectiveness of our model. As depicted in Figures~\ref{fig:good_case_2}, \ref{fig:good_case_3}, and \ref{fig:bad_case_1}, these case studies illustrate both the strengths and limitations of the proposed cybernetic loop.

Figure~\ref{fig:good_case_2} illustrates the model’s successful identification of abnormal events in a complex video scene. Through iterative reasoning and re-examination, the model accurately detects an artillery attack, demonstrating the ability of the cybernetic loop to refine attention and correct initial misinterpretations.

In Figure~\ref{fig:good_case_3}, the model's capability in object counting under dynamic conditions is showcased. Within a fast-paced VR game setting, \name{} correctly counts four cartoon pillows to the right of a blue lightsaber. This highlights the model’s strength in spatial reasoning and its capacity to integrate visual and textual cues through feedback mechanisms, effectively avoiding common counting errors.

Figure~\ref{fig:bad_case_1} presents a failure case in which the model misidentifies a missing item in a refrigerator. Due to incomplete or imprecise key frames selection by the Sensor module, the second round of inference draws an incorrect conclusion, falsely reporting a missing “potato” instead of the actual “hamburger.” This case shows a limitation of our method: inaccurate key frames extraction may, in certain instances, hinder the effectiveness of second-round visual enhancement, thereby failing to support meaningful self-correction.

\section{Limitations and Future Work Discussion}
\label{sec:lim_future}
While \name{} demonstrates notable improvements in test-time video understanding, several limitations remain. First, the current key frames extraction relies on attention drift over video and subtitle segments. Although some critical frames are often covered, this approach may \textbf{introduce noisy or irrelevant frames}. Our cybernetic loop can mitigate their impact through selective correction, but more principled methods for noise filtering, temporal search and the utilization of signals remain important directions for future work. 

Second, \textbf{current state-of-the-art MLLMs exhibit limited capacity for temporally grounded reasoning}--that is, the ability to precisely align and integrate the information from visual frames, subtitles, and questions along the temporal axis during the reasoning process. As a result, although \name{} yields significant gains on knowledge-centric benchmarks--which rely more on symbolic reasoning, logical inference, and mathematical derivation, capabilities that existing MLLMs are relatively better equipped to scale--its improvements on perceptual-heavy benchmarks are less pronounced. We believe that combining \name{} with future MLLMs possessing stronger multi-modal temporal grounded reasoning capabilities may yield greater benefits.

Another limitation lies in \textbf{inference efficiency}. As the number of inference paths ($N$) and iterations increases, test-time latency grows manyfold. Developing more efficient implementations of the cybernetic loop, potentially via strategy pruning, presents another valuable avenue for future research. 

Overall, while \name{} opens up a novel perspective on test-time adaptive reasoning, its full potential can be further unlocked through improvements in both base model capability and control system efficiency.

% \section{Rationale}
% \label{sec:rationale}
% % 
% Having the supplementary compiled together with the main paper means that:
% % 
% \begin{itemize}
% \item The supplementary can back-reference sections of the main paper, for example, we can refer to \cref{sec:intro};
% \item The main paper can forward reference sub-sections within the supplementary explicitly (e.g. referring to a particular experiment); 
% \item When submitted to arXiv, the supplementary will already included at the end of the paper.
% \end{itemize}

\end{document}